\documentclass[12pt]{article}
\usepackage{amsmath}

\usepackage{mathtools}
\usepackage{graphicx}
\usepackage{color}
\usepackage{natbib}
\usepackage{setspace}
\usepackage{amsfonts}
\usepackage{authblk}
\usepackage{url}

\usepackage[top=1.5in, bottom=1in, left=1in, right=1in]{geometry}

\bibpunct{(}{)}{;}{a}{}{,} 

\title{Interactive Graphics for Visually Diagnosing Forest Classifiers in R}
\author[1]{Natalia da Silva }
\author[2]{ Dianne Cook}
\author[3]{Eun-Kyung Lee}

\affil[1]{Department of Statistics, Iowa State University}
\affil[2]{Department of Econometrics and 
Business Statistics, Monash University}
\affil[3]{Department of Statistics, Ewha Womans University}
\date{April 2017}
\begin{document}
\maketitle 

\begin{abstract}
This paper describes structuring data and constructing plots to explore forest classification models interactively. A forest classifier is an example of an ensemble, produced by bagging multiple trees. The process of bagging and combining results from multiple trees, produces numerous diagnostics which, with interactive graphics, can provide a lot of insight into class structure in high dimensions. Various aspects are explored in this paper, to assess model complexity, individual model contributions, variable importance and dimension reduction, and uncertainty in prediction associated with individual observations. The ideas are applied to the random forest algorithm, and to the projection pursuit forest, but could be more broadly applied to other bagged ensembles. Interactive graphics are built in \textbf{R}, using the \textbf{ggplot2}, \textbf{plotly}, and \textbf{shiny} packages.
\end{abstract}

\section{Introduction}
The random forest (RF) algorithm~\citep{breiman1996bagging} was one of the first ensemble classifiers developed. It combines the predictions from individual classification and regression trees (CART)~\citep{breiman1984cl}, built by bagging observations~\citep{breiman1996bagging}. It also samples variables at each tree node. These produce diagnostics in the form of uncertainty in predictions for each observation, importance of variables for the prediction, predictive error for future samples based on out-of-bag (OOB) case predictions, and similarity of observations based on how often they group together in the trees.

Ensemble classifiers have grown in popularity~\citep{dietterish00}~\citep{talbot09}, and the basic ideas behind the random forest can be applied to virtually any type of model. The benefits for classification are reduced variability in predictive error, and the suite of diagnostics provides the potential for better understanding the class structure in the high-dimensional data space. The use of visualization on these diagnostics, in association with multivariate data plots, completes the process to support a better understanding of the underlying problem.

A conceptual framework for model visualization can be summarized in three strategies: (1) visualize the model in the data space, (2) look all members of a collection of a model and (3) explore the complete process of model fitting~\citep{wickham2015visualizing}. The first strategy is to explore how well the model captures the data characteristics (model in the data space), which contrasts determining if the model assumptions hold (data in the model space). The second strategy is to look at a group of models instead of only the best. This strategy can offer a broad understanding of the problem by comparing and contrasting possible models. The last strategy focuses on the exploration of the process of the model fit in addition to the end result.

There has been some, but not a lot of, research on visualizing classification models.
\cite{urbanek2002exploring} presents interactive tree visualization implemented in the java software called \textbf{KLIMT} that include zooming, selection, multiple views, interactive pruning and tree construction as well as the interactive analysis of forests of trees using treemaps. \cite{cutler15raft} developed a java package called \textbf{RAFT} to visualize a forest classifier, that included variable selection, parallel coordinate plots, heat maps and scatter plots of some diagnostics. Linking between plots is limited. \cite{quach2012interactive} presents interactive forest visualization using the R package \textbf{iPlots eXtreme} \citep{urbanek2011iplots}, where several displays are shown in the one window with some linking between them available. \cite{Silva2016} describes visualizing components of an ensemble classifier.

This paper describes structuring interactive graphics to facilitate visual exploration of ensemble classifiers, using RFs and projection pursuit forests (PPF)~\citep{dasilvappforest} as examples. The PPF algorithm builds on the projection pursuit tree (PPtree)~\citep{lee2013pptree} algorithm, which uses projection pursuit at each tree node to find the best linear combination of variables to separate the classes. The visualization approach is consistent with the framework in \cite{wickham2015visualizing}, and the implementation is built on the newest interactive graphics available in \textbf{R}. The purpose is to provide readily available tools for users to explore and improve ensemble fits, and obtain an intuition for the underlying class structure in data. Interactive plots are a key component for model visualization that help the user see multivariate relationships and be more efficient in the model diagnosis. Multiple levels of data are constructed for exploration: observation, model and ensemble summaries.

The paper is structured as follows.  Section \ref{key} describes the ensemble components to be accessed.  Section \ref{visen} maps the ensemble components to the visual elements. The web app is described in \ref{app} and further work is discussed in section \ref{fur}.

\newpage

\section{Diagnostics in forest classifiers}\label{key}

The diagnostics typically available are:

\begin{itemize} \itemsep 0in
  \item {\em Out-of-bag error:} For each model, in the ensemble, some cases of the original data are not used. Predicting the response for these cases gives a better estimate for the error of the model with future data. The OOB error rate is a measure for each model that is combined in the ensemble, and is used to provide the overall error of the ensemble.
  \item {\em Uncertainty measure for each observation:} Across individual (classification) models we can compute the proportion of times that a case is predicted to be each class. If a case is always predicted to be the true class, there is no uncertainty about an observation. Cases that are proportionately predicted to be multiple classes indicate difficult to classify observations. They may be important by indicating neighborhoods of the data space that would benefit from a more complex model, or more simply, they may be errors in measurements in the data.
  \item {\em Variable importance:} Each model uses samples of variables. With this, the accuracy of the models can be compared when the variable is included or omitted. There are several versions of statistics that use this to provide a measure of the variable importance for prediction.
  \item {\em Similarity measure for pairs of observations:} In each model, each pair of observations will be either in the same terminal node or not. This is used to compute a proximity matrix.  Cluster analysis on this matrix  can be used to follow up the classification to assess the original labeling. It may suggest improvements or mistakes in original labels.
\end{itemize}

In addition to these overall ensemble statistics, each component model has its own diagnostics, measuring error, variables utilized, and class predictions. Visualization will enable the individual models to be examined, relate these to the data and to their contribution to the ensemble.

\section{Mapping ensemble diagnostics to visual components}~\label{visen}

This section describes the mapping of diagnostics to visualizations. These are illustrated using the Australian crabs data~\citep{CM74}. The data has 200 cases, 5 predictors and 4 classes (combinations of species and sex, blue male, blue female, orange male and orange female). The predictors are: FL (the size of the frontal lobe length, in mm), RW (rear width, in mm), CL (length of mid-line of the carapace, in mm), CW (maximum width of carapace, in mm), BD (depth of the body; for females, measured after displacement of the abdomen, in mm). This is old data but it provides a good illustration of the visual methods.

\subsection{Individual models: PPtree}

\begin{figure}[hbpt]
\centering
\includegraphics[scale=.7]{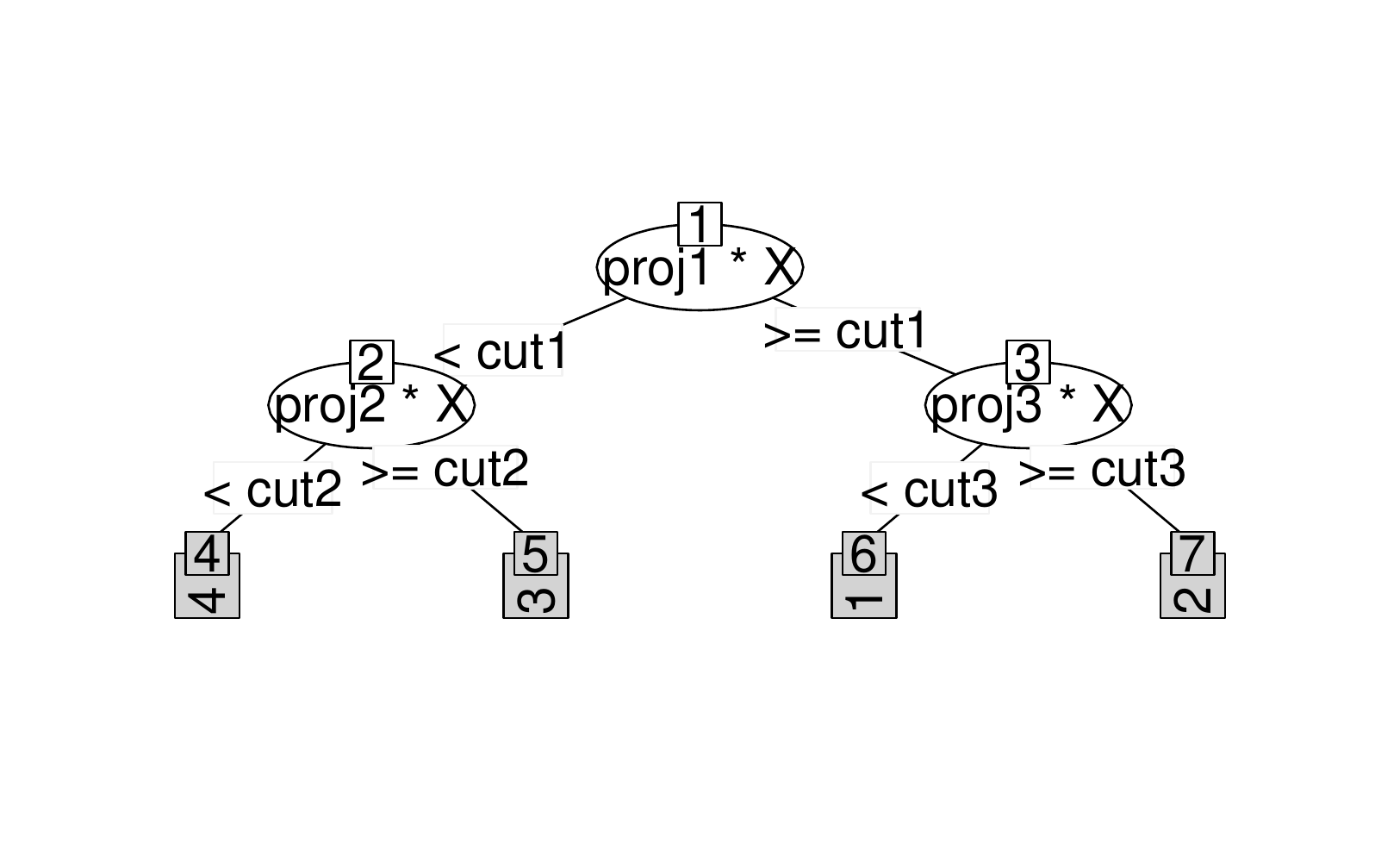} 
\includegraphics[scale=.9]{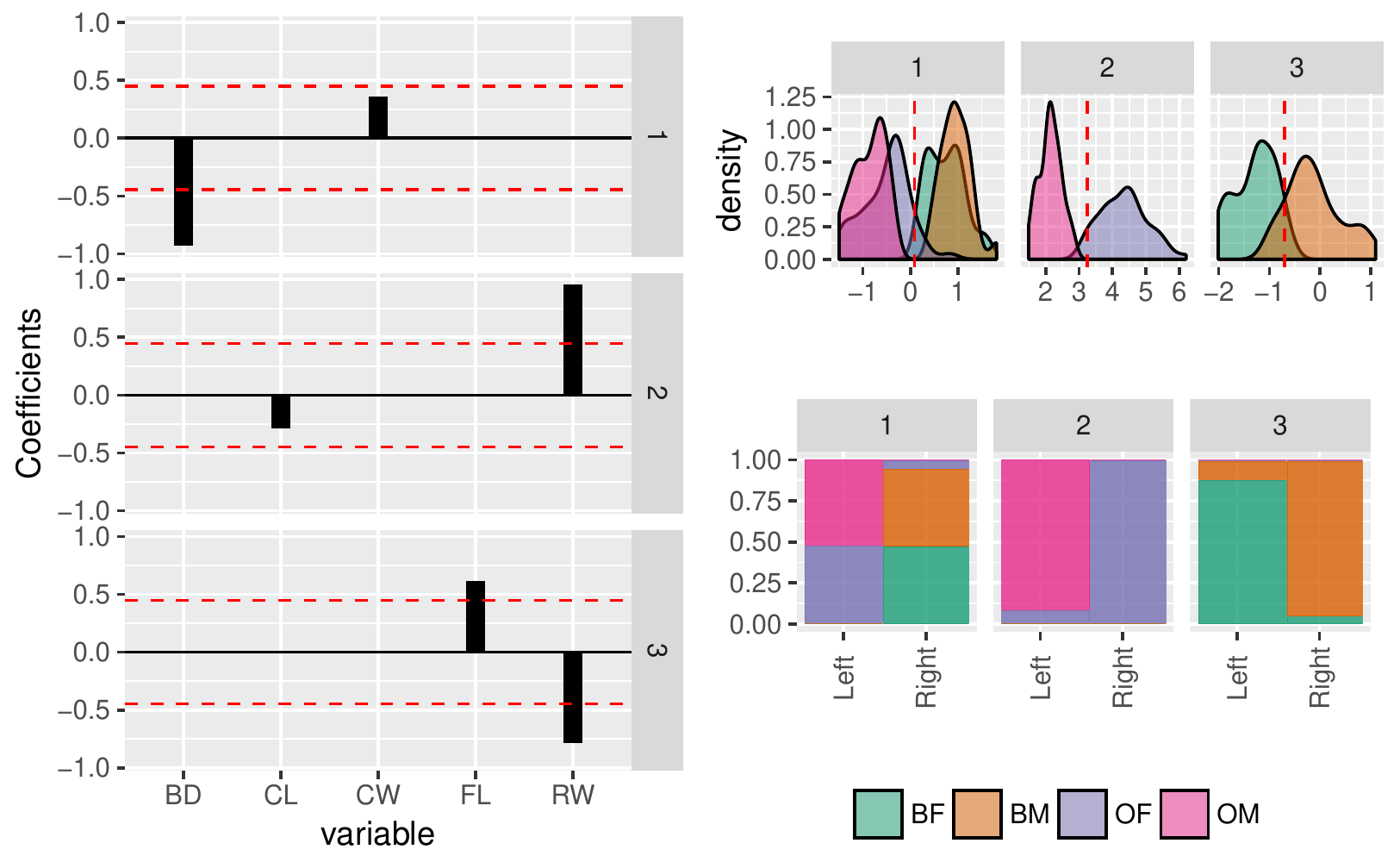} 
\caption{Visualizing the PPtree model of the crab data. The tree has three nodes (top). The density plots show the data projections at each node, colored by group (middle). The dashed vertical red line indicates the split value of each node. At node 1 the blue species is separated from orange species. Nodes 2 and 3 separate the sexes, which are more confused for the blue species. Mosaic plots of the confusion table for each split (bottom). Node 1 shows the clear split of the species, with a small number of misclassifications. Node 2 where orange females are separated from orange males indicates small number of misclassifications. Node 3 where blue females are separated from blue males, shows a larger misclassification than for orange specie.  
    \label{pptree}}
\end{figure}

The PPF is composed of individual projection pursuit trees. Figure \ref{pptree} shows a visual ensemble of plots of a tree model on the crab data. There are three nodes for the four class problem. The nodes of this tree are based on projections of the data, the coefficients of which form the building block to calculate the variable importance. The density plot displays the data projection at each node, and the mosaic plot shows the confusion matrix for the nodes.
The package \textbf{PPtreeViz} provides visual tools to diagnose a PPtree model. The PPF builds on these, and modified a little.   The PPtree model is simpler than a regular classification tree, because the classes are mostly separated by combinations of variables -- just three projections are needed to see the differences between the four classes.

\subsection{Variable importance}

The PPF algorithm calculates variable importance in two ways: (1) permuted importance using accuracy,  and (2) importance based on projection coefficients of standardized variables.

The permuted variable importance is comparable with the measure defined in the classical random forest algorithm. It is computed using the OOB sample for the tree $k\;\;(B^{(k)})$ for each $X_j$ predictor variable.  Then the
permuted importance of the variable $X_j$ in the tree $k$ can be defined as:

\[
IMP^{(k)}(X_j) = \frac{\sum_{i \in B^{(k)} } I(y_i=\hat y_i^{(k)})-I(y_i=\hat y_{i,P_j}^{(k)})}{|B^{(k)}|}
\]

\noindent where $\hat y_i^{(k)}$
is the predicted class for observation $i$ in tree $k$ and $y_{i,P_j}^{(k)}$ is the predicted class for observation $i$ in tree $k$ after permuting the values for variable $X_j$. The global permuted importance measure is the average importance over all the trees in the forest. 
This measure is based on comparing the accuracy of classifying OOB observations, using the true class with permuted (nonsense) class.

For the second importance measure, the coefficients of each projection are examined. The magnitude of these values indicates importance, if the variables have been standardized. The variable importance for a single tree is computed by a weighted sum of the absolute values of the coefficients across nodes. The weights takes the number of classes in each node into account~\citep{lee2013pptree}.
Then the importance of the variable $X_j$ in the PPtree $k$ can be defined as:

\[
  IMP_{pptree}^{(k)}(X_j)=\sum_{nd = 1}^{nn}\frac{|\alpha_{nd}^{(k)}|}{cl_{nd} }
\]

Where $\alpha_{nd}^{(k)}$ is the projected coefficient for node $nd$, variable $k$, and $nn$ the total number of node partitions in the tree $k$.

The global variable importance in a PPforest then can be defined in different ways. The most intuitive is the average variable importance from each PPtree across all the trees in the forest.

\[
IMP_{ppforest1}(X_j)=\frac{\sum_{k=1}^K IMP_{pptree}^{(k)}(X_j)}{K}
\]
Alternatively we have defined a global importance measure for the forest as a weighted mean of the absolute value of the projection coefficients across all nodes in every tree. The weights are based on the projection pursuit indexes in each node ($Ix_{nd}$), and 1-(OOB-error of each tree)($acc_k$).

\[IMP_{ppforest2}(X_j)=\frac{\sum_{k=1}^K acc_k \sum_{nd = 1}^{nn}\frac{Ix_{nd}|\alpha_{nd}^{(k)}|}{nn }}{K}
\]

\begin{figure}[hbpt]
\centering
\includegraphics[scale=.9]{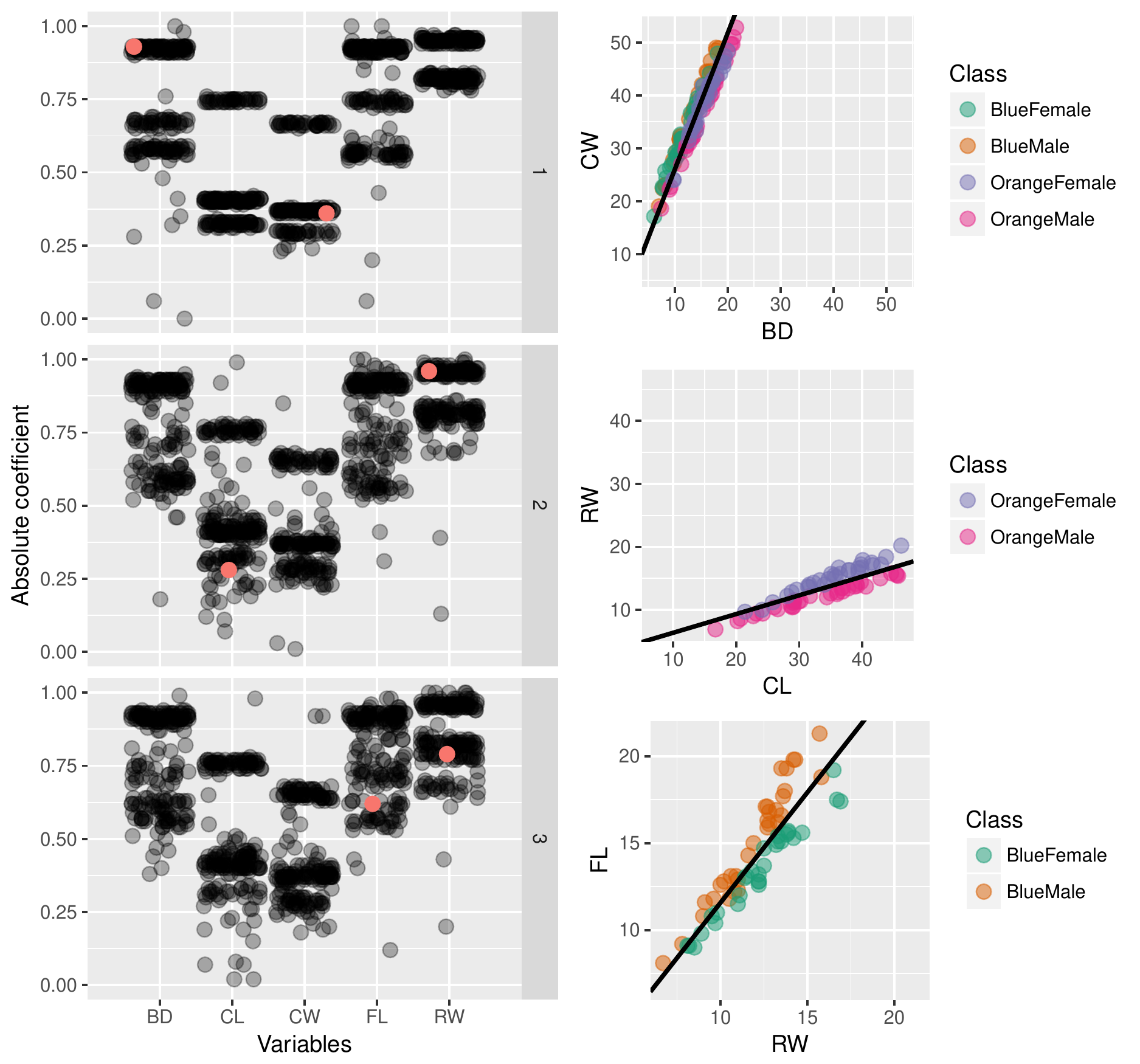} 
  \caption{Visualising the variable importance of all the trees in the forest, for three nodes. Each node of each tree has an importance value for each variable. The values for the whole forest are displayed using a side-by-side jittered dot plot. The importance values are the absolute values of the projection coefficients. The red points correspond to these values for the tree shown in Figure \ref{pptree}.  Two variables are randomly selected at each node for creating the best projection, and split. The plots are right show the variables used and the split made at each of the nodes of this tree.  \label{impotree}}
\end{figure}

Figure \ref{impotree} shows the absolute projection coefficient of the top three nodes for all the trees in a forest model. This information is displayed by a side-by-side jittered dot plot. The red dots correspond to the absolute coefficient values for the tree model of Figure \ref{pptree}. The forest was built using random samples of two variables for each node, hence there are two coefficients for each node. At node 1, BD has a high value and CW contributes much less. The scatterplot at right shows these two variables and the resulting boundary between groups that this would produce. Node 2 uses CL and RW, and RW contributes the most to the separation. The plot at right shows the boundary that is induced. Node 3 uses FL and RW, and this is a much more even contribution by the two variables.
For each tree in the forest different decision rules are defined, the resulting boundaries on the previous plots are based on Rule 1 $= \frac{m_1}{2} + \frac{m_2}{2} $, where $m_1$ and $m_2$ are the mean of the left and right groups at each node.

\subsection{Similarity of cases}

For each tree, every pair of observations can be in the same terminal node or not. Tallying this up across all trees in a forest gives the proximity matrix, an $n\times n$ matrix of the proportion of trees that the pair share a terminal node. It can be considered to be a similarity matrix.

\begin{figure}[hbpt]
\centering
\includegraphics[scale=.9]{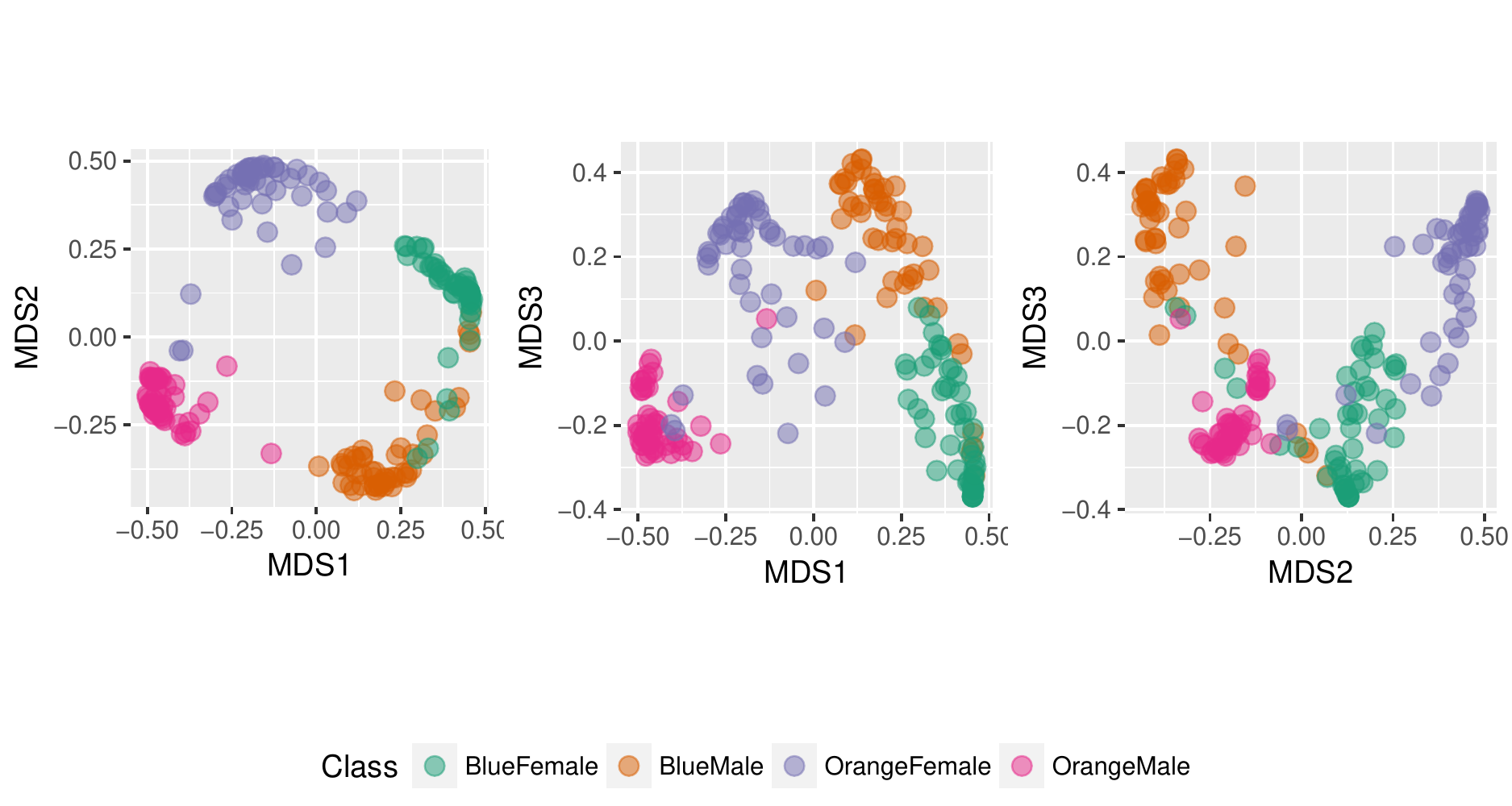} 
  \caption{Examining similarity between cases, using pairwise plots of multidimensional scaling into 3D. It can be seen that most cases are grouped closely with their class, and particularly that the two species are distinct. There is more confusion of cases between the sexes. \label{prox1}}
\end{figure}

Multidimensional scaling (MDS) is used to reduce the dimension of this matrix, to view the similarity between observations. MDS transforms the data set into a low-dimensional space where the distances are approximately the same as in the full $n$ dimensions. With $G$ groups, the low-dimensional space should be no more than $G-1$ dimensions. Figure \ref{prox1} shows the MDS plots for the 3D space induced by the four groups of the crab data. Color indicates the true species and sex. For this data two dimensions are enough to see the four groups separated quite well. Some crabs are clearly more similar to a different group, though, especially in examining the sex differences.

\subsection{Uncertainty of cases}

The vote matrix ($n \times p$) contains the proportion of times each observation was classified to each class, while oob. Two approaches to visualize the vote matrix information are used.

A ternary plot is a triangular diagram used to display compositional data with three components. More generally, compositional data can have any number of components, say $p$, and hence is constrained to a $(p-1)$-D simplex in $p$-space. The vote matrix is an example of compositional data, with $G$ components.

With G classes the ternary plot idea is generalized to a $(G-1)-D$ simplex~\citep{sutherland2000orca,schloerke}. This is one of the approaches used to visualize the vote matrix.

For the crab data, $G=4$ and the generalized ternary plot will be a tetrahedron. The  \textbf{tourr} package~\citep{wickham2011tourr} can be used to view it (e.g. \url{https://vimeo.com/170522736}).

\begin{figure}[hbpt]
\centering
\includegraphics[scale=.9]{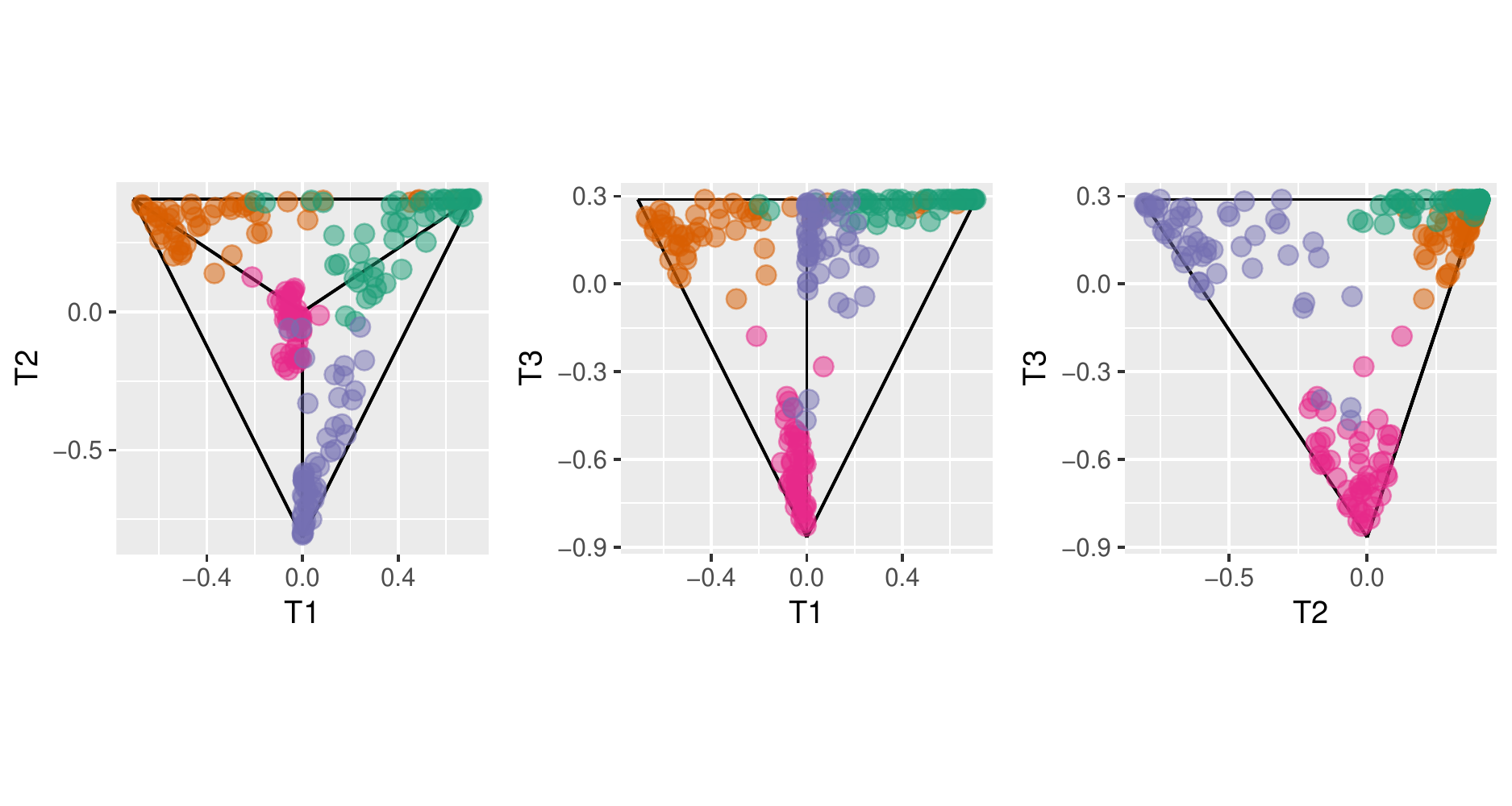} 
  \caption{Generalized ternary plot ((G-1)-D simplex, here it is a tetrahedron) representation of the vote matrix for four classes. The tetrahedron is shown pairwise. Each point corresponds to one observation and color is the true class. This is close but not a perfect classification, since the colors are concentrated in the corners and there are some mixed colors in each corner.}
\label{tetra}
\end{figure}

Figure \ref{tetra} shows the tetrahedron structure for the crab vote matrix shown in three pairwise views. With well-separated classes, the colored points will each concentrate into one of the vertices. This is close but not perfect, indicating some crabs are commonly incorrectly predicted.

Because visualizing the vote matrix with a $(G-1)$-D tetrahedron requires dynamic graphics, a low-dimensional option is also provided. For each class, each case has a value between 0-1. A side-by-side jittered dotplot is used for the display, where class is displayed on one axis and proportion is displayed on the other. For each dotplot, the ideal arrangement is that points are concentrated at 0 or 1, and only at 1 for their true class. This data is close to the ideal but not perfect, e.g. there are a few blue male crabs (orange) that are frequently predicted to be blue females (green), and a few blue female crabs predicted to be another class.

\begin{figure}[hbpt]
\centering
\includegraphics[scale=.9]{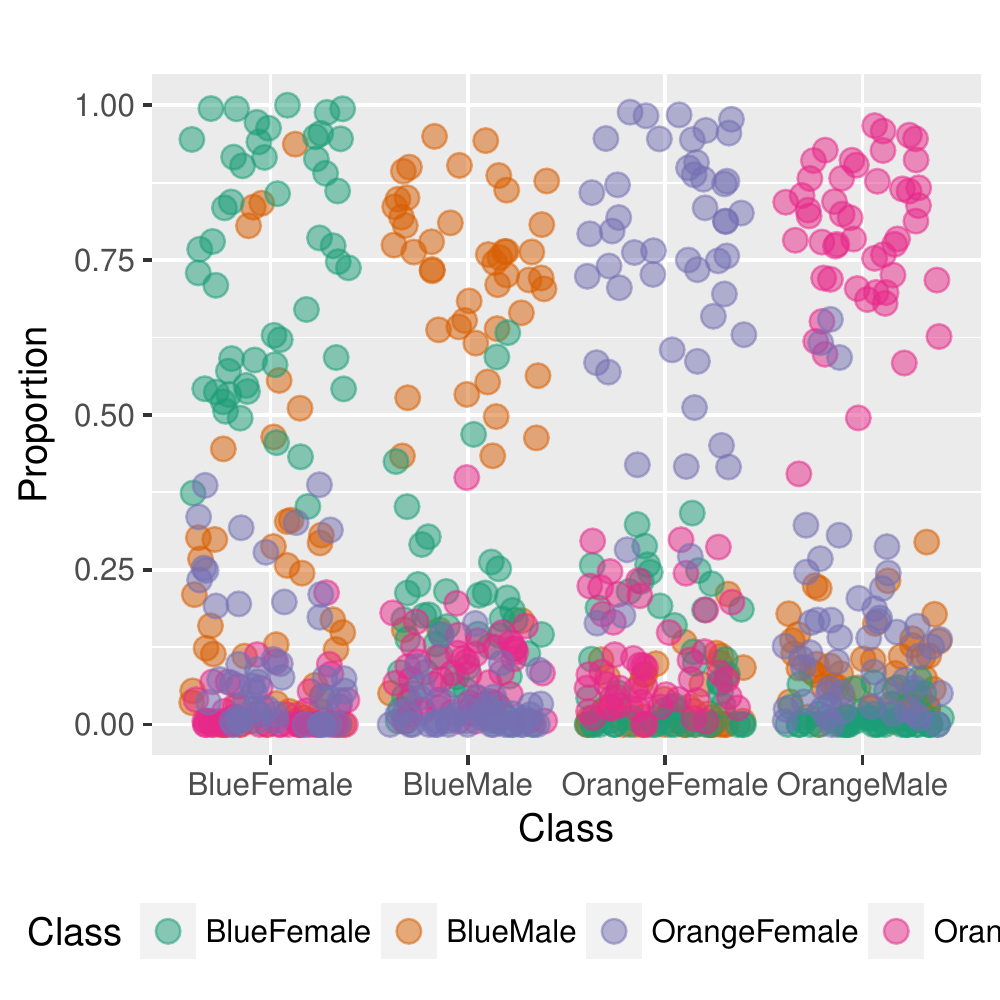} 
\caption{Another representation of the vote matrix, as a jittered side-by-side dotplot. It is not as elegant as the ternary plot, but it is useful because it places focus on each group. Each dotplot shows the proportion of times the case was predicted into the group, with 1 indicating that the case was always predicted to the group and 0 being never. On each dotplot, a single color dominates the top, indicating fairly clear distinctions between classes. Crabs from the blue species, both male and female, have more uncertainty in predictions, as seen by more crabs from other classes having higher vote proportions, than is seen in the orange species.}
\label{sideby}
\end{figure}

\newpage

\section{Interactive web app }\label{app}

Interaction is added to the plots described in Section \ref{visen} and other plots, and they are organized into an interactive web app using \textbf{shiny}~\citep{chang11shiny} for exploring the ensemble model. The app is organized into three tabs, individual cases, models, and performance comparison, to provide a model diagnostic tool. Interaction is provided as mouse-over labeling, mouse-click selection, and brushing, with results linked across multiple plots. The app takes advantage of new tools provided in the \textbf{plotly}~\citep{plotly} package, developed as a part of Sievert's PhD thesis research~\citep{sievertthesis}.

The \textbf{plotly} functions directly translate a static \textbf{ggplot2}  object by  extracting  information  and  storing it in JavaScript Object Notation (JSON). This information is passed as input to a javascript function, to produce a web graphic.  Interactions in a single display, and links between different graphics, are two key tasks an interactive visualization should accomplish~\citep{xie2014reactive}.

As \cite{sievertthesis} describes one of the biggest difficulties for the app in order manage linking between plots is the data structure management for each widget. Each widget has it own data structure and interaction. Putting them into the structure of a shiny app facilitates access to the widget data, and coordinates selections across multiple plots.

The fishcatch data~\citep{fishcatch} is used to illustrate the shiny app characteristics. It contains 159 observations, with 6 physical measurement variables, and  7 types of fish, all caught from the same lake (Laengelmavesi) near Tampere in Finland. There are 35 bream, 11 parkki, 56 perch, 17 pike, 20 roach, 14 smelt and 6 whitewish. The shiny app showing fishcatch data can be accessed at \url{https://natydasilva.shinyapps.io/shinyppforest}.

\subsection{Individual cases}

\begin{figure}[hbpt]
\centering
\includegraphics[scale=.7]{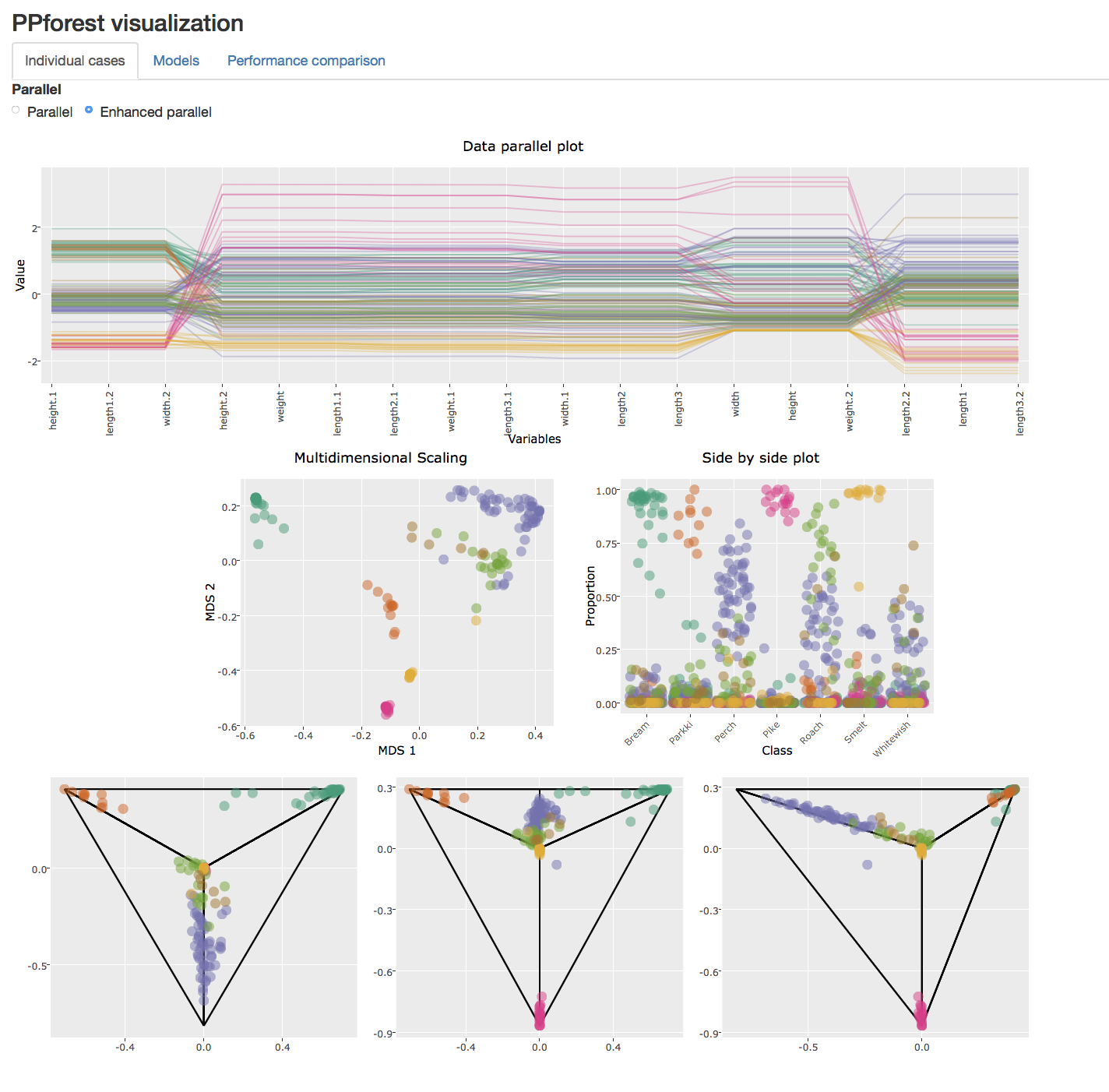}
\caption{Entry page for the web app, focusing on examining cases, in terms of similarity and uncertainty. The top plot shows the data, the remaining plots show similarity of cases and uncertainty in predictions. All of the plots are linked, so that selecting elements of one will generate changes in the others. \label{tab1}}
\end{figure}

This tab is designed to examine uncertainty in the classification of observations, and also to explore the similarity between pairs of observations. The data feeding the display is an $n\times p$ data frame, containing the original data, and the model statistics generated from the full $n\times G$ vote matrix, along with its generalized ternary coordinates, and the first two MDS projections of the proximity matrix.
Figure \ref{tab1} shows the arrangement of plots.
The plots in the tab are (1) a parallel coordinate plot (PCP) of the data, (2) the MDS display of the proximity matrix, (3) side-by-side jittered dotplot and (4) generalized ternary plot of the vote matrix. Each of these plots are interactive in the sense that each one presents individual interactions (mouse-over) and they are linked so that selections in one display are propagated to other plots (clicking and selecting).

This selection of plots enables aspects of the model, relating to performance for individual cases, to be examined in the data space. The data plot is an essential elements following the {\em model-in-the-data-space} philosophy of \citet{wickham2015visualizing}. The choice was made to use a parallel coordinate plot because it provides a space-efficient display of the data. Alternatives include the tour, a dynamic plot, or a scatterplot matrix. Theoretically, either of these could be substituted or added.

\begin{figure}[hbpt]
\centering
\includegraphics[scale=.5]{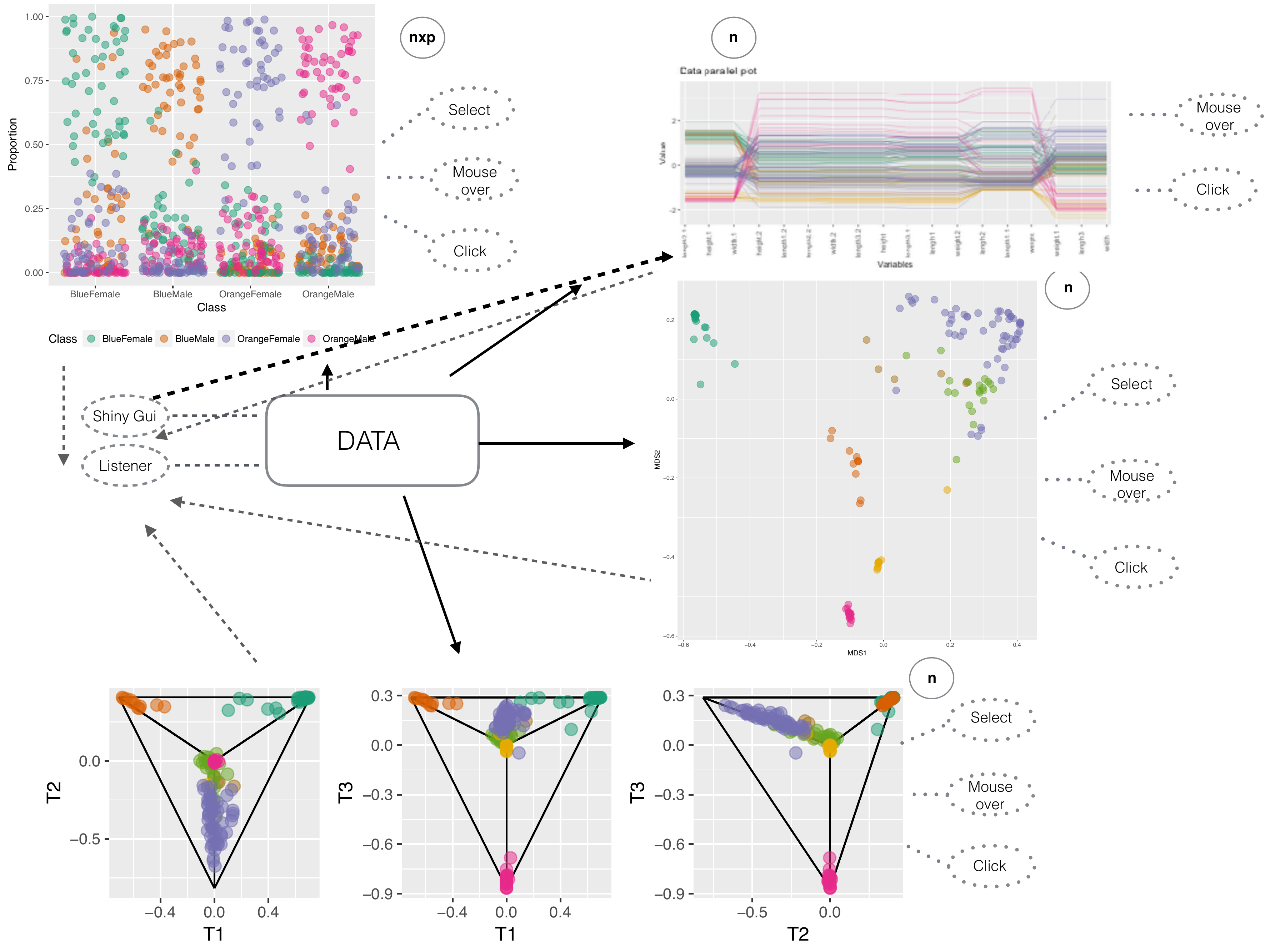}
\caption{Schematic diagram illustrating the interactivity in and between plots for individual level exploration panel of the web app. From the data, the four plots are generated. Each plot has a listener attached which collects user actions. When a user selects a point or line in any of the displays, it makes a change in the data which propagates updates to each of the plots.} 
\label{tab1diag}
\end{figure}

The diagram in Figure \ref{tab1diag} illustrates the data pipeline~\citep{BAHM88,wickham2009plumbing} for the interactive graphics in the case level tab. Solid lines indicate notifications from the source data to the plots, and dashed lines indicate notification of user action on the plot, that notifies the data source of actions to take. The data table is a reactive object, that has a listener associated with it. Each of the plots is reactive, and has numerous listeners. When users make selections on a plot, either by clicking or group selection, a change to the data is made in terms of an update on the selected cases. This invokes a note to other plots to re-draw themselves. The linking between plots is effectively one-to-one, based on the row id of the data. The side-by-side jittered dotplot has $n\times p$ points, but selection can only be done within a dotplot. Selecting in one of the dotplots notifies the data table of the selection which triggers a re-draw of the other dotplots. Mouseovers on the plot pull additional information about the point or line under the cursor but doesn't link between plots.

Two alternatives can be selected in shiny to draw the parallel coordinate plot: parallel or enhanced. Parallel draws the classic PCP and enhanced draws a modified version where variables are repeated \citep{hurley2011eulerian}.  Because reading a PCP is really only possible for neighboring variables, the variables are repeated so that all variables are neighboring.

\newpage

\subsection{Models}

This second tab in the app focuses on teasing apart the forest to examine the qualities of each tree. For each tree, information on the variable importance, the projections used at each node, and the OOB error is available.  The data feeding into this tab is a list of models, along with the original data frame.
The tree id is displayed when we mouse over the jittered side-by-side plot. This information is useful because, based on the accuracy some trees could be pruned from the forest outside of the app.

\begin{figure}[hbpt]
\centering
\includegraphics[scale=.7]{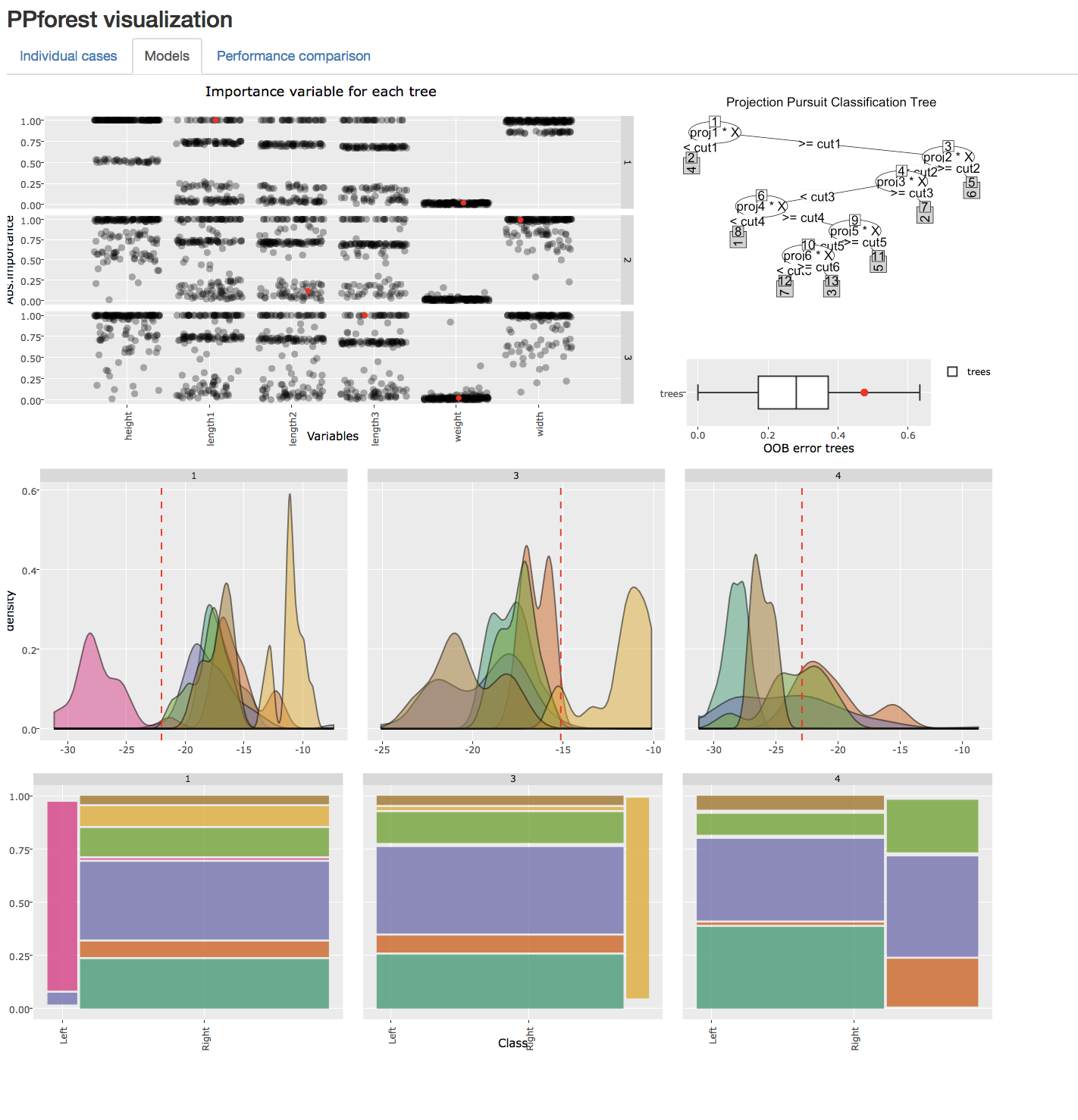}
\caption{The individual model tab in the web app. Variable importance is displayed as jittered dot plots for three nodes of all trees. This is linked to a display of the PPTree, a boxplot of the error for all trees in the forest, and display of the data showing splits at each of three nodes and confusion tables as mosaic plots. Clicking a point in the jittered dotplot triggers various updates: each of the importance values for the same tree are highlighted (red), the tree that this corresponds to is drawn, the error for the tree is shown on the boxplot (in red), and the data displays are updated to show the tree. \label{tab2}}
\end{figure}

Figure \ref{tab2} is a screenshot of the models tab. There are five plots, with varying levels of interaction: (1) a jittered side-by-side dotplot showing variable importance for the top three nodes of all trees in the forest, (2) a static display of one tree, (3) a boxplot of OOB error for all trees, (4) a faceted density plot of projected data at each node of the tree, with split point indicated by a vertical line, and (5) a mosaic plot showing the confusion matrix for each node of the tree.  The interaction is driven from the variable importance plot -- when the user selects a point in that display, the corresponding tree, density displays and mosaic plots are drawn. The tree plot from the \textbf{PPtreeViz} is used to visualize the selected tree structure. Also highlighted are the variable importance values for each variable for each of the top three nodes, and the OOB error value for the tree on the boxplot.

\begin{figure}[hbpt]
\centering
\includegraphics[scale=.5]{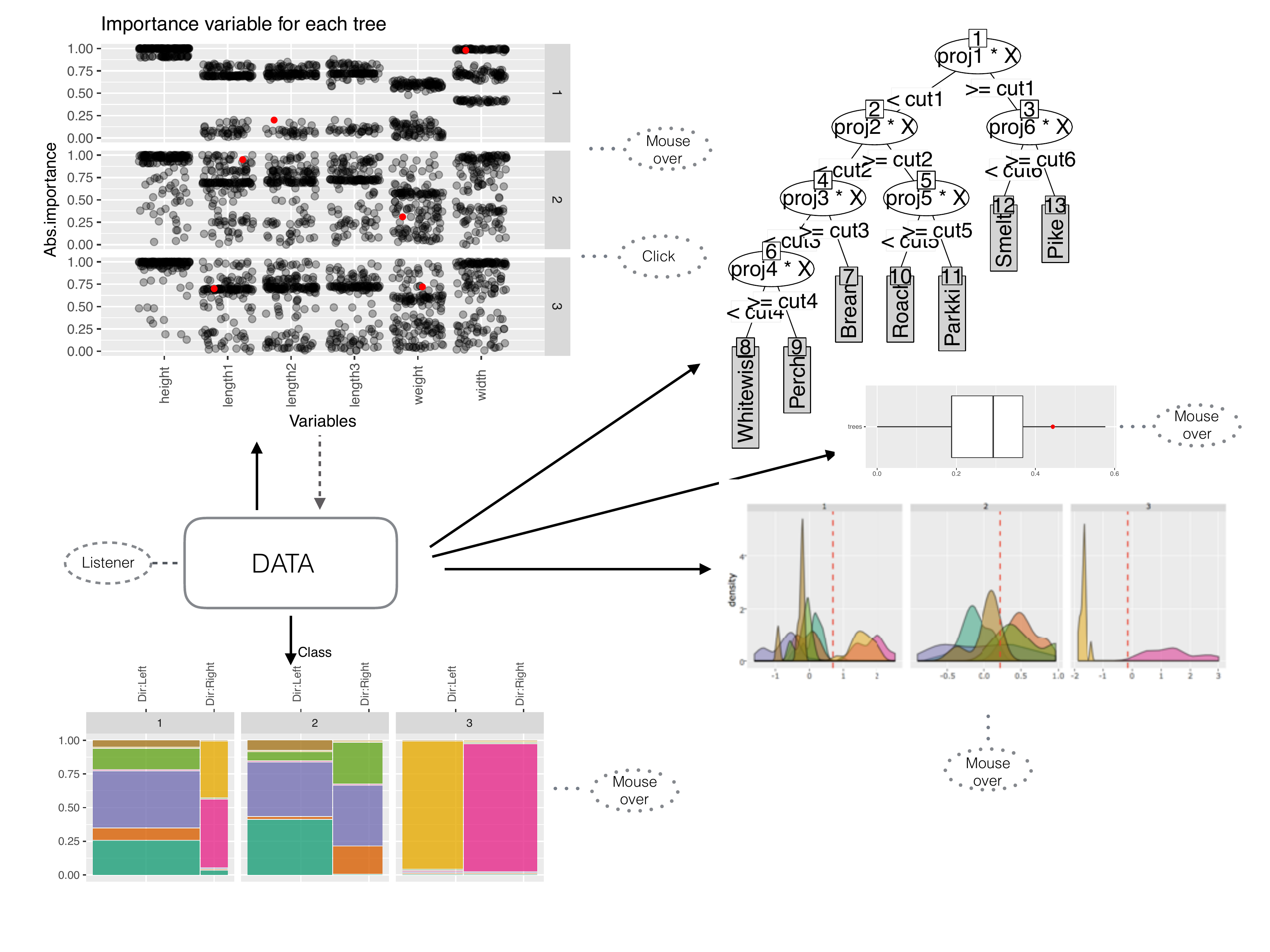}
\caption{Schematic diagram illustrating the interactivity in and between plots for model level exploration panel of the web app. Only the dotplot of variable importance is has click selection, which invokes changes of the tree display, boxplot, density plots and mosaic plots. Selecting a point, makes a change in the data, which propagates the importance values for other variables in this tree to be highlighted (red), draws the tree, highlights the error value of the tree, and shows the projections and confusion matrix for the three top nodes.}
\label{tab2diag}
\end{figure}

The diagram in Figure~\ref{tab2diag} illustrates the data pipeline for interactive graphics.
The data source is a \textbf{PPforest} object. Interaction is driven by the variable importance plot. Selecting a point triggers a change in the data, which cascades to re-draws of the other displays. Each plot has some information available on mouse over.

\subsection{Performance comparison}

\begin{figure}[hbpt]
\includegraphics[scale=.7]{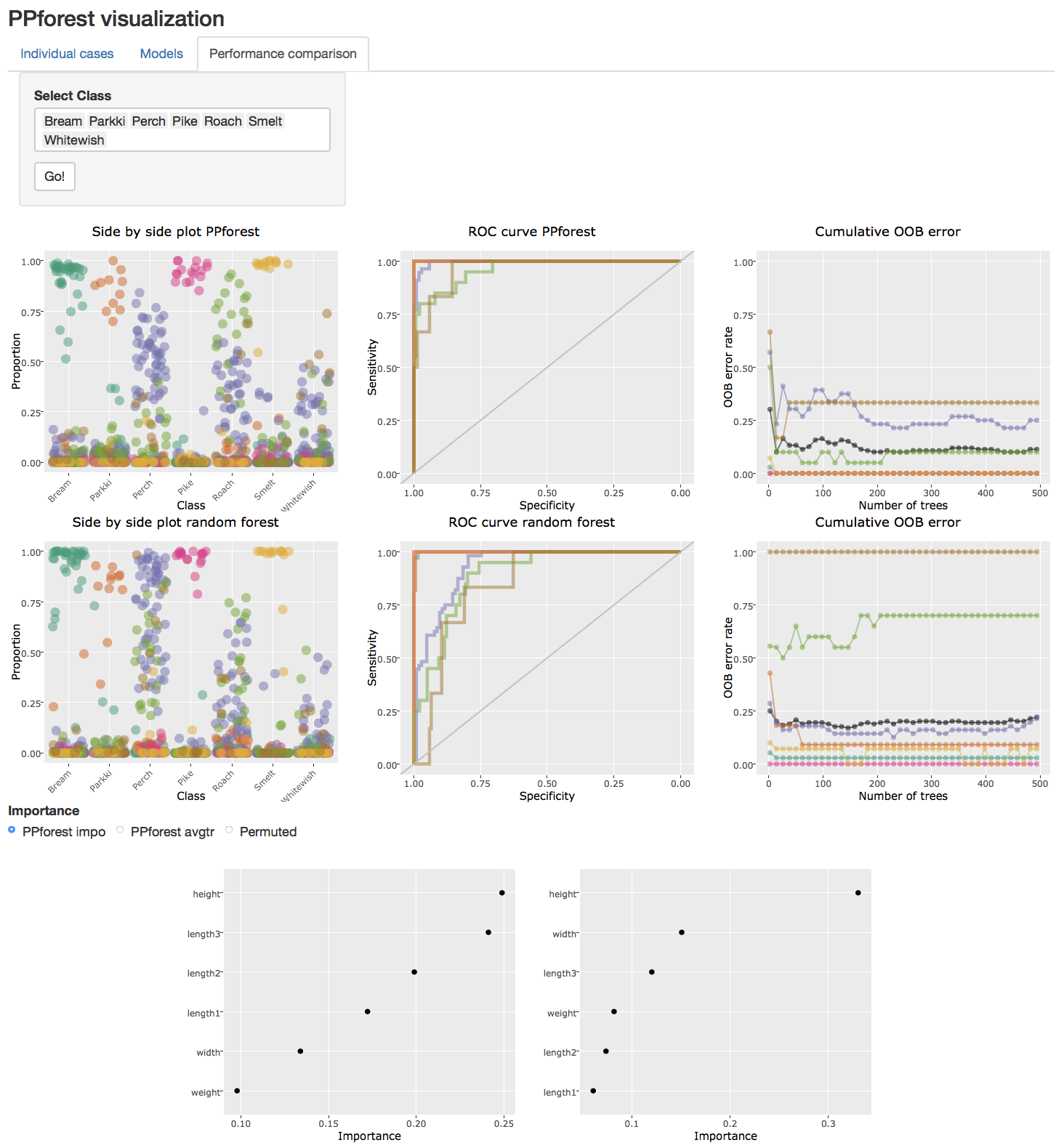}
\caption{Performance comparison tab of the web app. ROC curves displaying sensitivity against specificity for the classes are shown, along with the OOB error by number of trees used to build the forest, and overall variable importance. Displays are shown for the PPF and RF, for comparison purposes. Users can select a class to focus on, using the text entry box. \label{tab3}}
\end{figure}

The third tab (Figure \ref{tab3}) examines the PPF fit, and compares the result with a RF fit. There are four displays for each type of model: (1) Variable importance for all trees in the forest (same as in the models tab), (2) an receiver operating characteristic curve (ROC) curve comparing sensitivity and specificity for each class, (3) OOB error by number of trees, to assess complexity, (4) overall variable importance. There is very little interaction on this tab. Users can select to focus on a subset of classes, or choose the importance measure to show. Being able to focus on class can help to better understand how well the model performs across classes, and can be especially useful for unbalanced data. Examining the OOB error by trees enables an assessment of how few trees might be used to provide an equally accurate prediction of future data.

The  ROC is used to summarize the trade-off between sensitivity and specificity. The plot shows the sensitivity and specificity when a parameter of classifier is varied \citep{trevor2011elements}. The specificity and sensitivity was computed with the \textbf{pROC} package.
If more than two classes are available a multi-class ROC analysis is needed. Several solutions have been proposed for multi-class ROC. Some of the proposed reduced the multi-class problem to a set of binary problems. The approach used for a multi-class ROC analysis in this paper is called one-against-all ~\citep{allwein2000reducing}.

\newpage

\section{Discussion}\label{fur}

Having better tools to open up black box models will provide for better understanding the data, the model strengths and weaknesses, and how the model performs for future data. This visualisation app provides a selection of interactive plots to diagnose PPF models. This shell could be used to make an app for other ensemble classifiers. The philosophy underlying the collection of displays is ``show the model in the data space''  explained in \citet{wickham2015visualizing}. It is not easy to do this, and to completely take this on would require plotting the model in the $p$-dimensional data space. In the simplest approach, as taken here, it means to link the model diagnostics to displays of the data. Then it is possible to probe and query, to obtain a better understanding, such as finding regions in the data that prove difficult to fit, and detract from the predictive accuracy, or that don't adhere to model assumptions.

The app is implemented with new technology for interactive graphics provided by the \textbf{plotly} package. It is one of the first uses of these new tools.

One challenge to use plotly is that  when layers with different data are created in a ggplot2,  it is difficult to  specify the unique keys required for linking with another plot.

There are many possible extensions to the app, that could help it to be a tool for model refinement: (1) Using the diagnostics to weed out under-performing models in the ensemble; (2) Identifying and boosting models that perform well, particularly if they do well for problematic subsets of the data; (3) Problematic cases could be removed, and ensembles re-fit; (4) Classes as a whole could be aggregated or re-organised as suggested by the model diagnostics, to produce a more effective hierarchical approach to the multiple class problem. Working within the R environment makes all of these desires available using command line outside the app, given the unique ids of models and cases can be exported from the app.

The app has helped to identify ways to improve the  PPtree algorithm, and consequently the PPF model. These especially apply to multiclass problems. Multiple splits for the same class would enable nonlinear classifications. Split criteria tend to place boundaries too close to some groups, due to heteroskedasticity being induced by aggregating classes. Forests are not always better than their constituent trees, and if the trees can be built better, the forest will provide stronger predictions.

\newpage

\bibliographystyle{asa}      
\bibliography{bibViz}

\begin{thebibliography}{26}
\newcommand{\enquote}[1]{``#1''}
\expandafter\ifx\csname natexlab\endcsname\relax\def\natexlab#1{#1}\fi

\bibitem[{Allwein et~al.(2000)Allwein, Schapire, and
  Singer}]{allwein2000reducing}
Allwein, E.~L., Schapire, R.~E., and Singer, Y. (2000), \enquote{Reducing
  multiclass to binary: A unifying approach for margin classifiers,}
  \textit{Journal of machine learning research}, 1, 113--141.

\bibitem[{Breiman(1996)}]{breiman1996bagging}
Breiman, L. (1996), \enquote{Bagging predictors,} \textit{Machine learning},
  24, 123--140.

\bibitem[{Breiman et~al.(1984)Breiman, Friedman, Stone, and
  Olshen}]{breiman1984cl}
Breiman, L., Friedman, J., Stone, C.~J., and Olshen, R.~A. (1984),
  \textit{Classification and regression trees}, CRC press.

\bibitem[{Buja et~al.(1988)Buja, Asimov, Hurley, and McDonald}]{BAHM88}
Buja, A., Asimov, D., Hurley, C., and McDonald, J.~A. (1988),
  \enquote{{E}lements of a viewing iipeline for data analysis,} in
  \textit{Dynamic graphics for statistics}, eds. Cleveland, W.~S. and McGill,
  M.~E., Monterey, CA: Wadsworth, pp. 277--308.

\bibitem[{Campbell and Mahon(1974)}]{CM74}
Campbell, N.~A. and Mahon, R.~J. (1974), \enquote{A multivariate study of
  variation in two species of rock crab of genus {\em Leptograpsus},}
  \textit{Australian Journal of Zoology}, 22, 417--425.

\bibitem[{Chang et~al.(2015)Chang, Cheng, Allaire, Xie, and
  McPherson}]{chang11shiny}
Chang, W., Cheng, J., Allaire, J., Xie, Y., and McPherson, J. (2015),
  \enquote{shiny: Web application framework for R, R package version 0.11,} .

\bibitem[{Cutler and Breiman(2011)}]{cutler15raft}
Cutler, A. and Breiman, L. (2011), \enquote{RAFT: Random forest tool,} .

\bibitem[{da~Silva et~al.(2017)da~Silva, Cook, and Lee}]{dasilvappforest}
da~Silva, N., Cook, D., and Lee, E.-K. (2017), \enquote{Projection pursuit
  classification random forest,}
  \url{https://github.com/natydasilva/PPforestpaper}.

\bibitem[{Dietterich(2000)}]{dietterish00}
Dietterich, T.~G. (2000), \textit{Ensemble methods in machine learning}, New
  York: Springer Verlag, pp. 1--15.

\bibitem[{Hastie et~al.(2011)Hastie, Tibshirani, and
  Friedman}]{trevor2011elements}
Hastie, T.~J., Tibshirani, R.~J., and Friedman, J.~H. (2011), \textit{The
  elements of statistical learning: data mining, inference, and prediction},
  Springer.

\bibitem[{Hurley and Oldford(2011)}]{hurley2011eulerian}
Hurley, C.~B. and Oldford, R. (2011), \enquote{Eulerian tour algorithms for
  data visualization and the PairViz package,} \textit{Computational
  Statistics}, 26, 613--633.

\bibitem[{Lee et~al.(2013)Lee, Cook, Park, Lee, et~al.}]{lee2013pptree}
Lee, Y.~D., Cook, D., Park, J.-w., Lee, E.-K., et~al. (2013),
  \enquote{{PPtree}: Projection pursuit classification tree,}
  \textit{Electronic Journal of Statistics}, 7, 1369--1386.

\bibitem[{Puranen(2017)}]{fishcatch}
Puranen, J. (2017), \enquote{Finland fish catch,}
  \url{https://ww2.amstat.org/publications/jse/jse_data_archive.htm}.

\bibitem[{Quach(2012)}]{quach2012interactive}
Quach, A.~T. (2012), \enquote{Interactive random forests plots,} .

\bibitem[{Schloerke et~al.(2017)Schloerke, Wickham, Cook, and
  Hofmann}]{schloerke}
Schloerke, B., Wickham, H., Cook, D., and Hofmann, H. (2017), \enquote{Escape
  from Boxland: Generating a library of high-dimensional geometric shapes,}
  \textit{The R Journal}, \url{https://journal.r-project.org/archive/accepted}.

\bibitem[{Sievert(2017)}]{sievertthesis}
Sievert, C. (2017), \enquote{Interfacing R with the web for accessible,
  portable, and contents interactive data science,} .

\bibitem[{Sievert et~al.(2017)Sievert, Parmer, Hocking, Chamberlain, Ram,
  Corvellec, and Despouy}]{plotly}
Sievert, C., Parmer, C., Hocking, T., Chamberlain, S., Ram, K., Corvellec, M.,
  and Despouy, P. (2017), \textit{plotly: Create interactive web-based graphs
  via plotly's API}, r package version 1.1.0.

\bibitem[{Silva and Ribeiro(2016)}]{Silva2016}
Silva, C. and Ribeiro, B. (2016), \textit{Visualization of individual ensemble
  classifier contributions}, Cham: Springer International Publishing, pp.
  633--642.

\bibitem[{Sutherland et~al.(2000)Sutherland, Rossini, Lumley, Lewin-Koh,
  Dickerson, Cox, and Cook}]{sutherland2000orca}
Sutherland, P., Rossini, A., Lumley, T., Lewin-Koh, N., Dickerson, J., Cox, Z.,
  and Cook, D. (2000), \enquote{Orca: A visualization toolkit for
  high-dimensional data,} \textit{Journal of Computational and Graphical
  Statistics}, 9, 509--529.

\bibitem[{Talbot et~al.(2009)Talbot, Lee, Kapoor, and Tan}]{talbot09}
Talbot, J., Lee, B., Kapoor, A., and Tan, D.~S. (2009),
  \enquote{EnsembleMatrix: Interactive visualization to support machine
  learning with multiple classifiers,} in \textit{Proceedings of the SIGCHI
  Conference on Human Factors in Computing Systems (CHI '09)}, New York, NY,
  USA: Association for Computing Machinery, pp. 1283--1292.

\bibitem[{Urbanek(2008)}]{urbanek2002exploring}
Urbanek, S. (2008), \enquote{Visualizing trees and forests,} in
  \textit{Handbook of Data Visualization}, eds. Chen, C., H{\"a}rdle, W., and
  Unwin, A., Springer, Springer Handbooks of Computational Statistics, chap.
  III.2, pp. 243--264.

\bibitem[{Urbanek(2011)}]{urbanek2011iplots}
--- (2011), \enquote{iPlots eXtreme: next-generation interactive graphics
  design and implementation of modern interactive graphics,}
  \textit{Computational Statistics}, 26, 381--393.

\bibitem[{Wickham et~al.(2015)Wickham, Cook, and
  Hofmann}]{wickham2015visualizing}
Wickham, H., Cook, D., and Hofmann, H. (2015), \enquote{Visualizing statistical
  models: Removing the blindfold,} \textit{Statistical Analysis and Data
  Mining: The ASA Data Science Journal}, 8, 203--225.

\bibitem[{Wickham et~al.(2011)Wickham, Cook, Hofmann, Buja,
  et~al.}]{wickham2011tourr}
Wickham, H., Cook, D., Hofmann, H., Buja, A., et~al. (2011), \enquote{tourr: An
  R package for exploring multivariate data with projections,} \textit{Journal
  of Statistical Software}, 40, 1--18.

\bibitem[{Wickham et~al.(2009)Wickham, Lawrence, Cook, Buja, Hofmann, and
  Swayne}]{wickham2009plumbing}
Wickham, H., Lawrence, M., Cook, D., Buja, A., Hofmann, H., and Swayne, D.~F.
  (2009), \enquote{The plumbing of interactive graphics,} \textit{Computational
  Statistics}, 24, 207--215.

\bibitem[{Xie et~al.(2014)Xie, Hofmann, Cheng, et~al.}]{xie2014reactive}
Xie, Y., Hofmann, H., Cheng, X., et~al. (2014), \enquote{Reactive programming
  for interactive graphics,} \textit{Statistical Science}, 29, 201--213.

\end{thebibliography}

\end{document}